\documentclass[sigconf,nonacm]{acmart}
\renewcommand\footnotetextcopyrightpermission[1]{}

\usepackage{latexsym}
\usepackage{pifont}

\usepackage{amssymb}
\usepackage{enumitem}
\newcommand{\cmark}{\ding{51}}
\newcommand{\xmark}{\ding{55}}
\usepackage[T1]{fontenc}
\usepackage[utf8]{inputenc}
\usepackage{zi4}

\usepackage{algorithm}
\usepackage{algorithmic}

\usepackage{graphicx}
\usepackage{amsmath}
\usepackage{xcolor}
\usepackage{booktabs}

\AtBeginDocument{%
  }

\begin{document}

\title{Uncertainty-Aware Web-Conditioned Scientific Fact-Checking}

\author{Ashwin Vinod}
\affiliation{%
  \institution{University of Texas at Austin}
  \city{Austin}
  \state{Texas}
  \country{USA}
}
\email{ashwinv@utexas.edu}

\author{Katrin Erk}
\affiliation{%
  \institution{University of Massachusetts Amherst}
  \city{Amherst}
  \state{Massachusetts}
  \country{USA}
}
\email{kerk@umass.edu}


\begin{abstract}
Scientific fact-checking is vital for assessing claims in specialized domains such as biomedicine and materials science, yet existing systems often hallucinate or apply inconsistent reasoning, especially when verifying technical, compositional claims against an evidence snippet under source and cost/latency constraints. We present a pipeline centered on \emph{atomic} predicate–argument decomposition and \emph{calibrated, uncertainty-gated} corroboration: atomic facts are aligned to local snippets via embeddings, verified by a compact evidence-grounded checker, and only facts with uncertain support trigger domain-restricted web search over authoritative sources. The system supports both binary and tri-valued classification where it predicts labels from {\textsc{Supported}, \textsc{Refuted}, \textsc{NEI}} for three-way tasks. We evaluate under two regimes, \textit{Context-Only} (no web) and \textit{Context+Web} (uncertainty-gated web corroboration); when retrieved evidence conflicts with the provided context, we abstain with \textsc{NEI} rather than overriding the context. On multiple benchmarks, our framework surpasses the strongest benchmarks. In our experiments, web corroboration was invoked for only a minority of atomic facts on average, indicating that external evidence is consulted selectively under calibrated uncertainty rather than routinely. Overall, coupling atomic granularity with calibrated, uncertainty-gated corroboration yields more interpretable and context-conditioned verification, making the approach well-suited to high-stakes, single-document settings that demand traceable rationales, predictable cost/latency, and conservative abstention under domain and evidence grounding constraints.
\end{abstract}


\maketitle

\section{Introduction}
Scientific fact-checking is crucial in high-stakes domains such as biomedicine, public health, and materials science. However, state-of-the-art language models still struggle to verify technical statements: they often rely on surface cues, fail to correctly interpret predicate argument structure in complex sentences, and hallucinate when local evidence is sparse. As a result, their judgments can be inconsistent and weakly justified. This creates a need for methods that can verify technical claims from small, potentially noisy snippets, provide compute-efficient and explainable verdicts, and consult trusted web sources only when local evidence is insufficient.

We present a modular, context-aware, evidence-grounded pipeline. Claims are decomposed into atomic predicate argument facts, aligned to the most relevant evidence spans, and evaluated by a lightweight verifier. When uncertainty is high, a controlled corroboration step queries authoritative domains; all signals are aggregated into an interpretable verdict. This couples symbolic decomposition with probabilistic verification and targeted retrieval, balancing precision and coverage.

\noindent Empirically, the pipeline yields consistent test-time gains. On \textbf{BIONLI-300}, atomic decomposition plus web search improves factual F\textsubscript{1} by \textbf{+6.0} (66.7\% vs.\ 60.7\%) and Balanced Accuracy by \textbf{+7.35} points over sentence-level verification with \textit{MiniCheck}. Uncertainty-triggered corroboration adds a further \textbf{+4.7} F\textsubscript{1} and raises \textbf{PubMedFact1k} Macro-F\textsubscript{1} by \textbf{+1.2} over the strongest benchmark, reducing ambiguous NEI$\leftrightarrow$(Supported/Refuted) cases. The same pipeline transfers to climate fact-checking, achieving strong performance on \textsc{CLIMATE-FEVER} under both \textsc{Context-Only} (text-only) and \textsc{Context+Web} (constrained corroboration) settings with identical thresholds and prompts. Because our system is built around the same compact verifier (MiniCheck) and differs only in \emph{structure} (atomic facts), \emph{selection} (local snippets), and \emph{confident search} (uncertainty-gated corroboration), these gains isolate the value of grounded verification rather than larger models, improving accuracy, interpretability, and cost.

We recommend our pipeline when: (i) \textbf{traceability and abstention matter} (per-atom rationales and \textsc{NEI} abstentions when web and local evidence disagree); (ii) \textbf{cost/latency is constrained} (a compact verifier in the inner loop and retrieval only for uncertain atoms); (iii) \textbf{source control is required} (web corroboration restricted to vetted domains such as NIH/WHO/CDC). In short, choose \textsc{Atomic+Search}\footnote{Code and data available at: \href{https://ashwinn-v.github.io/uncertfc/}{https://ashwinn-v.github.io/uncertfc/}} when you need \emph{explanations and predictable cost} under domain and provenance constraints.

To support further research, we release \textsc{PubMedFact1k}, a three-way (\textsc{Supported}/\textsc{Refuted}/\textsc{NEI}) medical claim verification dataset derived from the \textsc{PubMedQA} dataset \cite{jin2019pubmedqa}. Together, these results show that combining atomic reasoning, calibrated uncertainty handling, and domain-aware corroboration enables more faithful and reproducible scientific fact-checking.

\section{Related Work}

Closed-book QA suggests that autoregressive transformers retain substantial factual knowledge, but whether such parametric memory alone suffices for veracity classification remains unclear. On \textit{SciFact}, zero-shot GPT-4 yield promising results with a few in-domain exemplars \citep{alvarez2024zero}, yet performance drops substantially on \textit{SciFact-Open} and related COVID/fake-news settings without retrieval \citep{wang2023explainable,faruk2024evaluating,huang2023fakegpt}, motivating retrieval with grounded checking. Prior approaches such as post-hoc revision (\textsc{RARR}, ``\emph{research}→\emph{revise}'') \citep{gao2022rarr}, broad tool-augmented checking (\textsc{FacTool}, Web/Scholar/Python) \citep{chern2023factool}, and RAG provenance verification via cross-encoder NLI \citep{sankararaman2024provenance} do not explicitly structure verification. In contrast, our \textsc{Atomic+Search} decomposes claims into $\leq$25-word predicate–argument units inspired by OpenIE, entailment graphs, and PropBank SRL \citep{etzioni2008open,levy2014focused,palmer2005proposition}; selects localized evidence windows; and verifies each atomic fact with a compact MiniCheck-style evidence-grounded model \citep{tang2024MiniCheck}, invoking domain-restricted web corroboration only under calibrated uncertainty. This atomic, uncertainty-gated loop unifies symbolic decomposition, probabilistic verification, and controlled corroboration in a single transparent pipeline, yielding finer-grained rationales and clearer failure modes (scope, negation, numerical qualifiers) than sentence-level verifiers and post-generation editors \citep{gao2022rarr,chern2023factool,sankararaman2024provenance}, as summarized in Table~\ref{tab:comp-ultra}.

\begin{table}[t]
\scriptsize
\setlength{\tabcolsep}{3pt}
\centering
\resizebox{\columnwidth}{!}{%
\begin{tabular}{lccccc}
\toprule
& \textbf{Atomic+Search} & \textbf{RARR} & \textbf{FacTool} & \textbf{Provenance} & \textbf{Base} \\
\midrule
\textbf{Explicit typed atomic facts}      & \cmark & \xmark & \xmark & \xmark & \xmark \\
\textbf{Uncertainty-gated retrieval}      & \cmark & \xmark & \xmark & \xmark & \xmark \\
\textbf{Domain-restricted external evidence} & \cmark & \xmark & \xmark & n/a   & n/a   \\
\textbf{Compact verifier (NLI)}           & \cmark & \xmark & \xmark & \cmark & \cmark \\
\textbf{LLM judge aggregation}            & \cmark & \xmark & \cmark & \xmark & \xmark \\
\textbf{Post-generation editing objective}& \xmark & \cmark & \xmark & \xmark & \xmark \\
\textbf{Tri-valued labels with \textsc{NEI}} & \cmark & \xmark & \xmark & \xmark & \cmark \\
\bottomrule
\end{tabular}%
}
\caption{Comparison of recent fact-checking methods. \cmark{} indicates the feature is present, \xmark{} indicates it is absent, and n/a means not applicable. “Base” represents a typical retrieve–verify pipeline.}
\label{tab:comp-ultra}
\end{table}

\section{Method}
\label{sec:method}

We propose an \emph{Atomic+Search} pipeline for scientific fact checking that couples (i) atomic fact decomposition, (ii) semantic local-evidence selection, (iii) lightweight verification with a grounded verifier, (iv) uncertainty-triggered, domain-constrained web corroboration, and (v) a final deductive judge. Given a natural-language claim $c$ and evidence $D$ (e.g., a paragraph or abstract), our system outputs a binary/ Tri-Valued verdict $\{\textsc{True},\textsc{False},\textsc{NEI}\}$ and a short rationale.

\subsection{Problem Formulation}
Let $c$ be a claim and $D$ the associated evidence context. We decompose $c$ into a set of atomic facts $\mathcal{F}=\{f_i\}_{i=1}^{m}$, where each $f_i$ targets a single predicate--argument tuple. For each $f_i$, a verifier returns a support probability $p_i \in [0,1]$ given evidence $D$:
\begin{equation}
p_i \;=\; \text{MiniCheck}(f_i \mid D). \label{eq:MiniCheck}
\end{equation}
Facts are labeled \textsc{supported}, \textsc{refuted} and \textsc{Uncertain} based on the grounding probabilities. Facts with $p_i \in [\alpha,\beta]$ (the \emph{Uncertain band}) trigger targeted web corroboration before a final verdict is produced.

Throughout, we adopt a \emph{single-document} verification regime: for each claim we assume exactly one evidence document $D$ is available. In our intended use cases, $D$ is either (i) the artifact supplied by the task, or (ii) the user's \emph{top retrieved evidence} the highest-ranked page/abstract they have already opened after a standard search. This isolates the contribution of atomic decomposition and calibrated verification from open-domain retrieval, mirrors how people assess claims while reading a single page, and avoids mixing potentially conflicting sources. When $D$ is insufficient, our uncertainty-gated corroboration step may consult additional sources; if those conflict with $D$, we abstain with \textsc{Uncertain} rather than overwrite the provided context.

\begin{figure}[t]
  \centering
  \includegraphics[width=.45\textwidth]{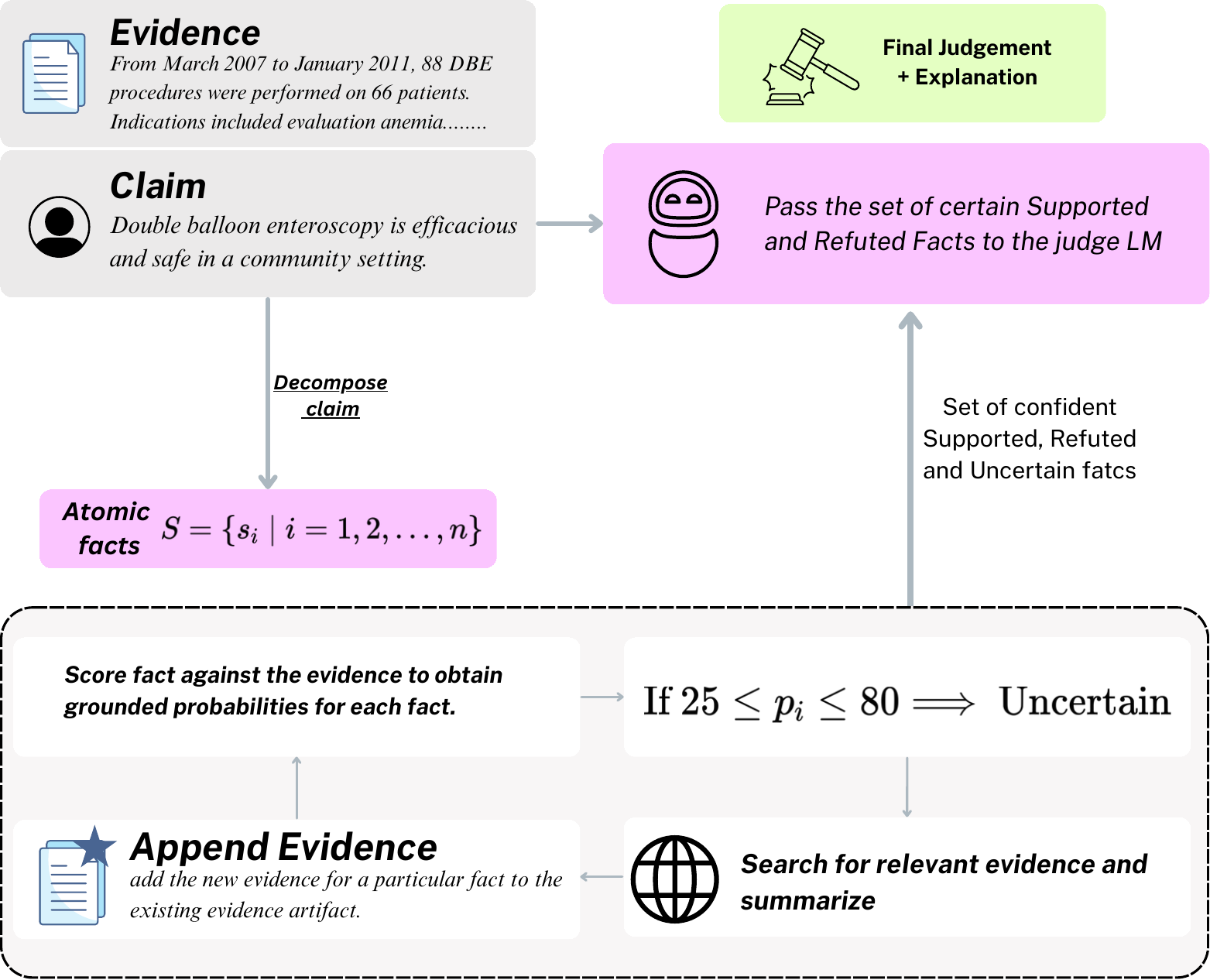}
  \caption{Atomic+Search pipeline}
  \label{fig:pipeline}
\end{figure}

\subsection{Datasets}

We evaluate our framework on three fact-checking datasets where each claim is paired with scientific evidence and a gold label, emphasizing grounded justification and traceable provenance. Unlike standard fact-checking benchmarks, scientific claim verification often requires decomposing claims and corroborating subclaims across multiple sources to establish true evidential support. \textbf{PubMedFact1k}\footnote{\url{https://huggingface.co/datasets/umbc-scify/PubmedFact1k}} is a 1{,}000 example biomedical claim verification set derived from the human-annotated PubMedQA PQA-Labeled split \citep{jin2019pubmedqa}, where each question is rewritten as a declarative claim paired with its abstract and \emph{yes/no/maybe} are mapped to \textsc{Supported}/\textsc{Refuted}/\textsc{NEI}.\textbf{BIONLI-300} is a 300-example subset of \textsc{BioNLI} that treats each hypothesis as the claim and its source abstract as context, mapping \emph{entailment} to \textsc{Supported} and \emph{contradiction} to \textsc{Refuted} \citep{bastan2022bionli}. \textbf{CLIMATE-FEVER} provides 1{,}535 climate-related claims, each linked to five Wikipedia evidence sentences; we merge the five sentences into a single document and evaluate only on the \textsc{Supported} and \textsc{Refuted} subsets \citep{diggelmann2020climate}.

\subsection{Uncertainty-Aware Web Corroboration}
\noindent Our pipeline (Fig.~\ref{fig:pipeline}) takes a claim \(c\) and document \(D\) and produces a structured, claim-level verdict through uncertainty-aware stages. First, a reasoning-capable LLM performs \textbf{atomic fact decomposition}, converting \((c, D)\) into short atomic facts \(\{f_i\}\) (each \(\leq 25\) words, expressing a single predicate--argument tuple) and returning strict JSON of the form \texttt{facts[\{id, text, targets\}]}. This yields consistent, fine-grained units for downstream verification.

Next, \textbf{semantic snippet selection} identifies local evidence for each fact. The document is chunked into overlapping \(\approx 420\)-character windows \(\{x_j\}\). For each fact \(f_i\), we form a query from its text and targets, compute embeddings for all queries and chunks using \texttt{text-embedding-3-large}, and select the window with highest cosine similarity; if embeddings are unavailable, we fall back to a deterministic token-overlap heuristic. Given fact--snippet pairs, a lightweight verifier, \textit{MiniCheck-7B}, produces a calibrated support probability for each fact. Facts with probability \(\geq 0.80\) are labeled \textsc{Supported}, those \(\leq 0.25\) as \textsc{Refuted}, using binary calibration instead of multi-class NLI to simplify aggregation and reduce label drift across biomedical domains.

\textbf{Uncertainty-triggered web corroboration} is applied only to facts with intermediate scores between 0.25 and 0.80. For these, we issue concise web queries that restate the fact and include its local snippet as context, restricting search to authoritative biomedical sources (e.g., PubMed, WHO, CDC, FDA, NIH, ClinicalTrials.gov, Wikipedia). Retrieved snippets are summarized with inline citations into an auxiliary text, which is concatenated with the snippet to form augmented evidence; MiniCheck-7B is re-run on this augmented evidence, and we record both the original and updated probabilities and whether rescoring occurred. The 0.25--0.80 band is chosen from a held-out calibration study of MiniCheck~\cite{pandit2025teaching}, where most ambiguous cases concentrate.

Finally, a \textbf{Judge LLM} aggregates fact-level outputs into a claim-level decision. It receives the original claim and document along with two fact sets after corroboration: \(\mathcal{S}\), facts labeled \textsc{Supported} (probability \(\geq 0.80\)), and \(\mathcal{R}\), facts labeled \textsc{Refuted} (probability \(\leq 0.25\)). Facts with intermediate probabilities are omitted for two-way decisions but reported as \textsc{Uncertain} in a three-way setting. Conditioned on these inputs, the Judge outputs strict JSON with fields \texttt{final\_verdict} (\textsc{Supported}, \textsc{Refuted}, or \textsc{NEI}), \texttt{explanation} (a short rationale referencing fact IDs), and \texttt{used\_facts}.

\subsection{Implementation Details}
We split $D$ into sentence groups with a maximum length of approximately 420 characters per chunk, trading off locality against coverage. For both decomposition and judging we use a chat LLM in JSON mode, while a high-capacity text-embedding model is employed for embeddings and \textit{MiniCheck-7B} is used for verification. We fix the high-confidence threshold at $\tau_{\text{hi}}{=}0.80$, define an uncertainty band of $[0.25, 0.80]$, and cap the maximum fact length at 25 words. Web-based corroboration is restricted to reputable scientific domains, including PubMed, WHO, CDC, FDA, NIH and  ClinicalTrials.gov, with the domain list configurable to match the target area (e.g., materials science). For each example we record the baseline verdict on $c$, per-fact probabilities before and after corroboration, the synthesized web evidence and citations, and the final verdict together with an explanation.

\section{Experimental Setup}
\label{sec:Experiments}

We evaluate \textsc{Atomic+Search} in two claim verification regimes: a binary setting (\textsc{Supported}/\textsc{Refuted}) and a three-way setting that additionally includes \textsc{NEI}. We compare against four families of approaches: 
(i) \textbf{compact verifiers and sentence-level baselines}: \textbf{MiniCheck} (sentence-level), a calibrated NLI-style checker; 
(ii) \textbf{closed-book and tool-augmented LLMs}: \textbf{GPT-5 Mini}, \textbf{Qwen-32B (Instruct/MAD)}, and \textbf{GPT-5 Mini + Search} (generic web-augmented prompting); 
(iii) \textbf{recent retrieval- and verification-centric systems}, \textbf{RARR} (retrieve–revise attribution editing),
and (iv) \textbf{reasoning-focused LMs}: \textbf{o1} (reasoning model with long-context verification). 
Our variants include the full \textbf{Atomic+Search} pipeline and ablations (Table~\ref{tab:ablate}). 
\begin{table*}[h]
\centering
\small
\setlength{\tabcolsep}{7pt}
\begin{tabular}{lccccccc}
\toprule
& \multicolumn{3}{c}{\textbf{BIONLI-300}} & \textbf{PubMedFact1k (3-way)} & \multicolumn{3}{c}{\textbf{CLIMATE-FEVER (2-way)}} \\
\cmidrule(lr){2-4}\cmidrule(lr){5-5}\cmidrule(lr){6-8}
\textbf{System} & Bal.\ Acc.\,$\uparrow$ & Rec.\,$\uparrow$ & F1\,$\uparrow$ & Macro F1\,$\uparrow$ & Bal.\ Acc.\,$\uparrow$ & Rec.\,(S)\,$\uparrow$ & F1\,(S)\,$\uparrow$ \\
\midrule
GPT-o1 & 65.8\% & 60.2\% & 64.9\% & 71.2\% & 69.70\% & 56.10\% & 65.05\% \\
MiniCheck & 61.35\% & 59.8\% & 60.7\% & --- & 69.10\% & 55.00\% & 64.00\% \\
GPT-5 Mini & 62.9\% & 58.7\% & 61.8\% & 68.5\% & 67.90\% & 54.20\% & 63.10\% \\
Qwen 32B MAD & 62.5\% & 59.1\% & 61.3\% & 61.8\% & 67.30\% & 53.60\% & 62.60\% \\
\midrule
RARR & 66.4\% & 62.3\% & 65.3\% & 72.3\% & 70.40\% & 57.80\% & 66.30\% \\
GPT-5 Mini + Search & 66.9\% & 62.7\% & 65.8\% & 72.5\% & 71.20\% & 58.50\% & 67.00\% \\
\textbf{Atomic+Search (ours)} & \textbf{68.7\%} & \textbf{65.4\%} & \textbf{66.7\%} & \textbf{73.7\%} & \textbf{73.83\%} & \textbf{62.14\%} & \textbf{70.04\%} \\
\bottomrule
\end{tabular}
\caption{\textbf{Main results.} Balanced Accuracy, Recall, and F1 on \textsc{BIONLI-300}; Macro-F\textsubscript{1} on \textsc{PubMedFact1k}; and Balanced Accuracy, Recall (Supports), and F1 (Supports) on \textsc{CLIMATE-FEVER}, evaluated only on the \textsc{Supported}/\textsc{Refuted} subsets.}
\label{tab:main}
\end{table*}

\subsection{Main Results}
\noindent Table~\ref{tab:main} reports results on \textbf{BIONLI-300} (2-way), \textbf{PubMedFact1k} (3-way), and \textbf{CLIMATE-FEVER} (2-way). On \textbf{BIONLI-300}, Our framework performs best overall, markedly improving Balanced Accuracy and F\textsubscript{1} over sentence-level \emph{MiniCheck} and tool-augmented LLM baselines; since \emph{MiniCheck} is our underlying verifier, these gains isolate the impact of atomic decomposition and \emph{confident search}. On \textbf{PubMedFact1k}, it achieves the top Macro-F\textsubscript{1}, outperforming \textbf{GPT-5 Mini + Search} and \textbf{GPT-o1}, and reducing borderline \textsc{NEI} vs.\ \textsc{Supported/Refuted} errors. On \textbf{CLIMATE-FEVER}, it again leads in Balanced Accuracy and F\textsubscript{1}, with consistent gains over both compact verifiers and larger LLMs, indicating strong out-of-domain generalization.

\subsection{Per component Importance}

\begin{table}[t]
\centering
\small
\begin{tabular}{lcc}
\toprule
\textbf{BIONLI-300} & F\textsubscript{1}\,$\uparrow$ & $\Delta$ vs.\ full \\
\midrule
Atomic+Search (full) & \textbf{66.7\%} & --- \\
\quad \textsc{No-Search} & 62.0\% & $-\,4.7$ \\
\quad \textsc{No-Atomic} & 60.3\% & $-\,6.4$ \\
\quad \textsc{MajorityVote (No-Judge)} & 52.1\% & $-\,14.6$ \\
\bottomrule
\end{tabular}
\caption{Ablations on \textsc{BIONLI-300}. Each line removes one component from the full pipeline.}
\label{tab:ablate}
\end{table}

On BIONLI-300 (Table~\ref{tab:ablate}), removing web corroboration (\textsc{No-Search}) reduces F1 from 66.7\% to 62.0\% ($-\,$\textbf{4.7}); removing atomic decomposition (\textsc{No-Atomic}) further drops F1 to 60.3\% ($-\,$\textbf{6.4}), indicating decomposition is the larger contributor. Replacing the Judge with majority vote (\textsc{MajorityVote}) yields 52.1\% F1 ($-\,$\textbf{14.6}), showing that calibrated fact-level aggregation is crucial. Overall, decomposition, corroboration, and principled aggregation are complementary.

\subsection{Retrieval and Corroboration Analysis}

\begin{figure}[t]
\centering
\includegraphics[width=0.60\linewidth]{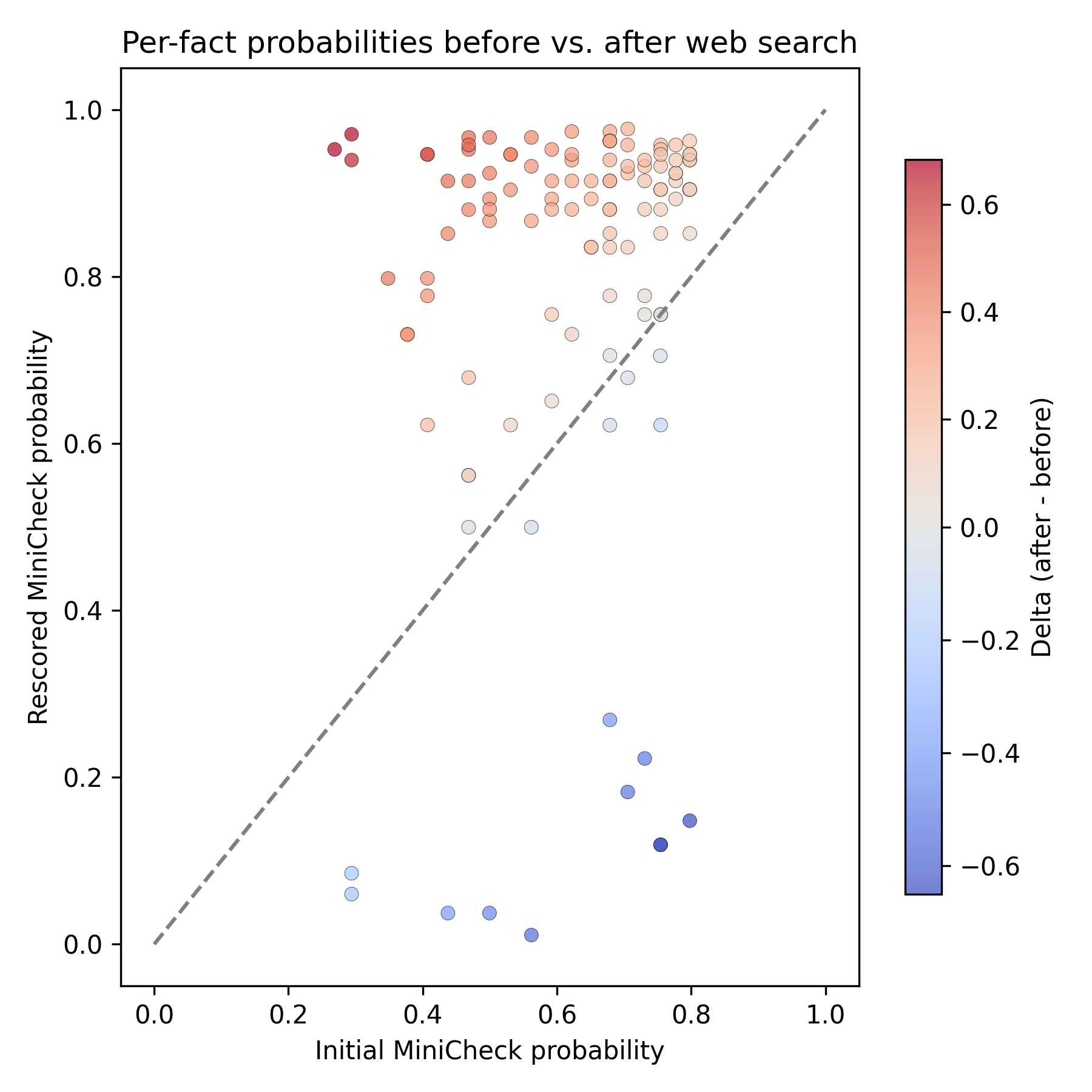}
\caption{Distribution of per-fact support $p_i$ before and after web corroboration on \textsc{BIONLI-300}.}
\label{fig:before-after}
\end{figure}

Figure~\ref{fig:before-after} plots per-fact support probabilities before and after corroboration on BIONLI-300. The distribution shifts rightward for many facts, reflecting successful evidence augmentation when local context is insufficient. Most of the helpful external evidence comes from authoritative medical sources (e.g., PubMed/NIH/NCBI), with Wikipedia contributing recall for broad biomedical background.

\section{Conclusion}
We introduced an uncertainty-aware, web-conditioned pipeline for scientific fact-checking at predicate–argument (``atomic'') granularity: claims are decomposed into minimal facts, each aligned to a local snippet, verified with a compact evidence-grounded checker, and, only under calibrated uncertainty, supplemented by domain-restricted corroboration before a final judge combines high-confidence atoms into an interpretable claim-level verdict. Across biomedical and out-of-domain evaluations, \textsc{Atomic+Search} consistently outperforms sentence-level verifiers, tool-augmented LLMs, and recent retrieval+verification benchmarks, while ablations show that atomic decomposition, uncertainty-gated corroboration, and the judge are complementary, corroboration is needed for only a minority of atoms, and the system abstains with \textsc{NEI} when retrieved evidence conflicts with the provided context (Tables~\ref{tab:main}--\ref{tab:ablate}). Our approach still relies on fixed calibration thresholds, curated domain allow-lists, and heuristic snippet selection. Future work includes adaptive calibration; tighter integration of decomposition, selection, and verification; richer temporal and quantitative reasoning; improved source-quality modeling; and human-in-the-loop workflows that surface atomic rationales. We hope that combining atomic structure with calibrated corroboration, together with our released components and \textsc{PubMedFact1k}, will spur more robust, context-aware scientific fact-checking.

\begin{acks}
This material is  based on research that is in part supported by DARPA for the SciFy program under agreement number HR00112520301.
\end{acks}

\bibliographystyle{ACM-Reference-Format}
\bibliography{custom}


\begin{thebibliography}{15}


\ifx \showCODEN    \undefined \def \showCODEN     #1{\unskip}     \fi
\ifx \showISBNx    \undefined \def \showISBNx     #1{\unskip}     \fi
\ifx \showISBNxiii \undefined \def \showISBNxiii  #1{\unskip}     \fi
\ifx \showISSN     \undefined \def \showISSN      #1{\unskip}     \fi
\ifx \showLCCN     \undefined \def \showLCCN      #1{\unskip}     \fi
\ifx \shownote     \undefined \def \shownote      #1{#1}          \fi
\ifx \showarticletitle \undefined \def \showarticletitle #1{#1}   \fi
\ifx \showURL      \undefined \def \showURL       {\relax}        \fi
\providecommand\bibfield[2]{#2}
\providecommand\bibinfo[2]{#2}
\providecommand\natexlab[1]{#1}
\providecommand\showeprint[2][]{arXiv:#2}

\bibitem[Alvarez et~al\mbox{.}(2024)]%
        {alvarez2024zero}
\bibfield{author}{\bibinfo{person}{Carlos Alvarez}, \bibinfo{person}{Maxwell Bennett}, {and} \bibinfo{person}{Lucy~Lu Wang}.} \bibinfo{year}{2024}\natexlab{}.
\newblock \showarticletitle{Zero-shot scientific claim verification using LLMs and citation text}. In \bibinfo{booktitle}{\emph{Proceedings of the Fourth Workshop on Scholarly Document Processing (SDP 2024)}}. \bibinfo{pages}{269--276}.
\newblock


\bibitem[Bastan et~al\mbox{.}(2022)]%
        {bastan2022bionli}
\bibfield{author}{\bibinfo{person}{Mohaddeseh Bastan}, \bibinfo{person}{Mihai Surdeanu}, {and} \bibinfo{person}{Niranjan Balasubramanian}.} \bibinfo{year}{2022}\natexlab{}.
\newblock \showarticletitle{BioNLI: Generating a Biomedical NLI Dataset Using Lexico-semantic Constraints for Adversarial Examples}.
\newblock \bibinfo{journal}{\emph{arXiv preprint arXiv:2210.14814}} (\bibinfo{year}{2022}).
\newblock


\bibitem[Chern et~al\mbox{.}(2023)]%
        {chern2023factool}
\bibfield{author}{\bibinfo{person}{I Chern}, \bibinfo{person}{Steffi Chern}, \bibinfo{person}{Shiqi Chen}, \bibinfo{person}{Weizhe Yuan}, \bibinfo{person}{Kehua Feng}, \bibinfo{person}{Chunting Zhou}, \bibinfo{person}{Junxian He}, \bibinfo{person}{Graham Neubig}, \bibinfo{person}{Pengfei Liu}, {et~al\mbox{.}}} \bibinfo{year}{2023}\natexlab{}.
\newblock \showarticletitle{FacTool: Factuality Detection in Generative AI--A Tool Augmented Framework for Multi-Task and Multi-Domain Scenarios}.
\newblock \bibinfo{journal}{\emph{arXiv preprint arXiv:2307.13528}} (\bibinfo{year}{2023}).
\newblock


\bibitem[Diggelmann et~al\mbox{.}(2020)]%
        {diggelmann2020climate}
\bibfield{author}{\bibinfo{person}{Thomas Diggelmann}, \bibinfo{person}{Jordan Boyd-Graber}, \bibinfo{person}{Jannis Bulian}, \bibinfo{person}{Massimiliano Ciaramita}, {and} \bibinfo{person}{Markus Leippold}.} \bibinfo{year}{2020}\natexlab{}.
\newblock \showarticletitle{Climate-fever: A dataset for verification of real-world climate claims}.
\newblock \bibinfo{journal}{\emph{arXiv preprint arXiv:2012.00614}} (\bibinfo{year}{2020}).
\newblock


\bibitem[Etzioni et~al\mbox{.}(2008)]%
        {etzioni2008open}
\bibfield{author}{\bibinfo{person}{Oren Etzioni}, \bibinfo{person}{Michele Banko}, \bibinfo{person}{Stephen Soderland}, {and} \bibinfo{person}{Daniel~S Weld}.} \bibinfo{year}{2008}\natexlab{}.
\newblock \showarticletitle{Open information extraction from the web}.
\newblock \bibinfo{journal}{\emph{Commun. ACM}} \bibinfo{volume}{51}, \bibinfo{number}{12} (\bibinfo{year}{2008}), \bibinfo{pages}{68--74}.
\newblock


\bibitem[Faruk(2024)]%
        {faruk2024evaluating}
\bibfield{author}{\bibinfo{person}{Tanjim~Bin Faruk}.} \bibinfo{year}{2024}\natexlab{}.
\newblock \showarticletitle{Evaluating the Performance of Large Language Models in Scientific Claim Detection and Classification}.
\newblock \bibinfo{journal}{\emph{arXiv preprint arXiv:2412.16486}} (\bibinfo{year}{2024}).
\newblock


\bibitem[Gao et~al\mbox{.}(2022)]%
        {gao2022rarr}
\bibfield{author}{\bibinfo{person}{Luyu Gao}, \bibinfo{person}{Zhuyun Dai}, \bibinfo{person}{Panupong Pasupat}, \bibinfo{person}{Anthony Chen}, \bibinfo{person}{Arun~Tejasvi Chaganty}, \bibinfo{person}{Yicheng Fan}, \bibinfo{person}{Vincent~Y Zhao}, \bibinfo{person}{Ni Lao}, \bibinfo{person}{Hongrae Lee}, \bibinfo{person}{Da-Cheng Juan}, {et~al\mbox{.}}} \bibinfo{year}{2022}\natexlab{}.
\newblock \showarticletitle{{RARR}: Researching and Revising What Language Models Say, Using Language Models}.
\newblock \bibinfo{journal}{\emph{arXiv preprint arXiv:2210.08726}} (\bibinfo{year}{2022}).
\newblock


\bibitem[Huang and Sun(2023)]%
        {huang2023fakegpt}
\bibfield{author}{\bibinfo{person}{Yue Huang} {and} \bibinfo{person}{Lichao Sun}.} \bibinfo{year}{2023}\natexlab{}.
\newblock \showarticletitle{FakeGPT: fake news generation, explanation and detection of large language models}.
\newblock \bibinfo{journal}{\emph{arXiv preprint arXiv:2310.05046}} (\bibinfo{year}{2023}).
\newblock


\bibitem[Jin et~al\mbox{.}(2019)]%
        {jin2019pubmedqa}
\bibfield{author}{\bibinfo{person}{Qiao Jin}, \bibinfo{person}{Bhuwan Dhingra}, \bibinfo{person}{Zhengping Liu}, \bibinfo{person}{William Cohen}, {and} \bibinfo{person}{Xinghua Lu}.} \bibinfo{year}{2019}\natexlab{}.
\newblock \showarticletitle{PubMedQA: A Dataset for Biomedical Research Question Answering}. In \bibinfo{booktitle}{\emph{Proceedings of the 2019 conference on empirical methods in natural language processing and the 9th international joint conference on natural language processing (EMNLP-IJCNLP)}}. \bibinfo{pages}{2567--2577}.
\newblock


\bibitem[Levy et~al\mbox{.}(2014)]%
        {levy2014focused}
\bibfield{author}{\bibinfo{person}{Omer Levy}, \bibinfo{person}{Ido Dagan}, {and} \bibinfo{person}{Jacob Goldberger}.} \bibinfo{year}{2014}\natexlab{}.
\newblock \showarticletitle{Focused Entailment Graphs for Open {IE} Propositions}. In \bibinfo{booktitle}{\emph{Proceedings of the Eighteenth Conference on Computational Natural Language Learning}}. \bibinfo{pages}{87--97}.
\newblock


\bibitem[Palmer et~al\mbox{.}(2005)]%
        {palmer2005proposition}
\bibfield{author}{\bibinfo{person}{Martha Palmer}, \bibinfo{person}{Daniel Gildea}, {and} \bibinfo{person}{Paul Kingsbury}.} \bibinfo{year}{2005}\natexlab{}.
\newblock \showarticletitle{The {P}roposition {B}ank: An Annotated Corpus of Semantic Roles}.
\newblock \bibinfo{journal}{\emph{Computational linguistics}} \bibinfo{volume}{31}, \bibinfo{number}{1} (\bibinfo{year}{2005}), \bibinfo{pages}{71--106}.
\newblock


\bibitem[Pandit et~al\mbox{.}(2025)]%
        {pandit2025teaching}
\bibfield{author}{\bibinfo{person}{Shrey Pandit}, \bibinfo{person}{Ashwin Vinod}, \bibinfo{person}{Liu Leqi}, {and} \bibinfo{person}{Ying Ding}.} \bibinfo{year}{2025}\natexlab{}.
\newblock \showarticletitle{Teaching with Lies: Curriculum DPO on Synthetic Negatives for Hallucination Detection}.
\newblock \bibinfo{journal}{\emph{arXiv preprint arXiv:2505.17558}} (\bibinfo{year}{2025}).
\newblock


\bibitem[Sankararaman et~al\mbox{.}(2024)]%
        {sankararaman2024provenance}
\bibfield{author}{\bibinfo{person}{Hithesh Sankararaman}, \bibinfo{person}{Mohammed~Nasheed Yasin}, \bibinfo{person}{Tanner Sorensen}, \bibinfo{person}{Alessandro Di~Bari}, {and} \bibinfo{person}{Andreas Stolcke}.} \bibinfo{year}{2024}\natexlab{}.
\newblock \showarticletitle{Provenance: A Light-weight Fact-checker for Retrieval Augmented LLM Generation Output}.
\newblock \bibinfo{journal}{\emph{arXiv preprint arXiv:2411.01022}} (\bibinfo{year}{2024}).
\newblock


\bibitem[Tang et~al\mbox{.}(2024)]%
        {tang2024MiniCheck}
\bibfield{author}{\bibinfo{person}{Liyan Tang}, \bibinfo{person}{Philippe Laban}, {and} \bibinfo{person}{Greg Durrett}.} \bibinfo{year}{2024}\natexlab{}.
\newblock \showarticletitle{Minicheck: Efficient fact-checking of llms on grounding documents}.
\newblock \bibinfo{journal}{\emph{arXiv preprint arXiv:2404.10774}} (\bibinfo{year}{2024}).
\newblock


\bibitem[Wang and Shu(2023)]%
        {wang2023explainable}
\bibfield{author}{\bibinfo{person}{Haoran Wang} {and} \bibinfo{person}{Kai Shu}.} \bibinfo{year}{2023}\natexlab{}.
\newblock \showarticletitle{Explainable Claim Verification via Knowledge-Grounded Reasoning with Large Language Models}.
\newblock \bibinfo{journal}{\emph{arXiv preprint arXiv:2310.05253}} (\bibinfo{year}{2023}).
\newblock


\end{thebibliography}

\end{document}